\pdfoutput=1

\documentclass[11pt]{article}

\usepackage{EMNLP2022}

\usepackage{times}
\usepackage{latexsym}

\usepackage[T1]{fontenc}

\usepackage[utf8]{inputenc}

\usepackage{microtype}

\usepackage{inconsolata}

\usepackage{multirow}
\usepackage{booktabs}
\usepackage{graphicx}
\usepackage{mathrsfs}
\usepackage{amsmath}
\usepackage{amsfonts}
\usepackage{bbm}
\usepackage[normalem]{ulem}
\usepackage[linecolor=orange]{todonotes}
\usepackage[multiple]{footmisc}

\DeclareMathOperator{\PLM}{PLM}
\DeclareMathOperator{\prob}{P}

\DeclareMathOperator{\TopK}{Top\textit{K}}

\DeclareMathOperator{\orig}{orig}
\DeclareMathOperator{\sym}{sym}

\title{SPE: Symmetrical Prompt Enhancement for Fact Probing}

\author{Yiyuan Li$^{*}$ \\
  UNC-Chapel Hill\\
  \texttt{yiyuanli@cs.unc.edu} \\\And
  Tong Che\thanks{$^*$ Equal contribution.} \,\thanks{\, Work was done at MILA.} \\
  NVIDIA\\
  \texttt{tongc@nvidia.com} \\\And
  Yezhen Wang\\
  Mila-Quebec AI Institute\\
  \texttt{yezhen.wang@mila.quebec}\\\AND
  Zhengbao Jiang\\
  Carnegie Mellon University\\
  \texttt{zhengbaj@cs.cmu.edu}\And
  Caiming Xiong\\
  Salesforce Research\\
  \texttt{cxiong@salesforce.com}\And
  Snigdha Chaturvedi\\
  UNC-Chapel Hill\\
  \texttt{snigdha@cs.unc.edu}
  }

\begin{document}
\maketitle
\begin{abstract}
Pretrained language models (PLMs) have been shown to accumulate factual knowledge during pretraining~\cite{petroni-etal-2019-language}. Recent works probe PLMs for the extent of this knowledge through prompts either in discrete or continuous forms. However, these methods do not consider symmetry of the task: object prediction and subject prediction. In this work, we propose Symmetrical Prompt Enhancement (SPE), a continuous prompt-based method for factual probing in PLMs that leverages the symmetry of the task by constructing symmetrical prompts for subject and object prediction. Our results on a popular factual probing dataset, LAMA,  show significant improvement of SPE over previous probing methods.
\end{abstract}

\section{Introduction}
Prompt-based learning proposes to formulate different NLP tasks into language modeling problems~\cite{schick-schutze-2021-exploiting}. It is a novel paradigm that effectively uses Pretrained Language Models (PLMs)~\cite{Liu2021PretrainPA}, and achieves comparable or better performance than fine-tuning \cite{lester-etal-2021-power}. 
Prompt-based learning has also been used for the task of factual knowledge probing in PLMs. In this task, the goal is to predict the (masked) object of factual tuples of type (subject, relation, object) using PLMs. Prompting methods assume that PLMs gather and store factual knowledge during their pre-training, and cloze-style prompts can be used to probe PLMs to gauge how much knowledge they contain~\cite{petroni-etal-2019-language}. The prompts are either handcrafted~\cite{petroni-etal-2019-language, Bouraoui2020InducingRK} or automatically generated~\cite{shin-etal-2020-autoprompt, haviv-etal-2021-bertese}. For example, to 
probe PLMs about their knowledge of geographic location of \textit{Luxembourg}, a \texttt{prompt} can be formed by filling \textit{Luxembourg} in the first blank of the following \texttt{template}: "\_\_\_\_ is located in \_\_\_\_.". An effective prompt will probe the PLM to output \textit{Europe} as the most likely prediction for the second blank. Such methods are promising but brittle. Minor changes in the template can lead to significant difference in the performance \cite{jiang-etal-2020-know}. Recent works have shown that continuous prompts obtained via gradient-based learning, are more effective and robust than discrete prompts since there are less restrictions on the search space ~\cite{Liu2021GPTUT, qin-eisner-2021-learning, zhong2021factual, Liu2021PretrainPA, newman2022padapters}.

\begin{figure}[t!]
	\centering
	\includegraphics[width=0.52\textwidth]{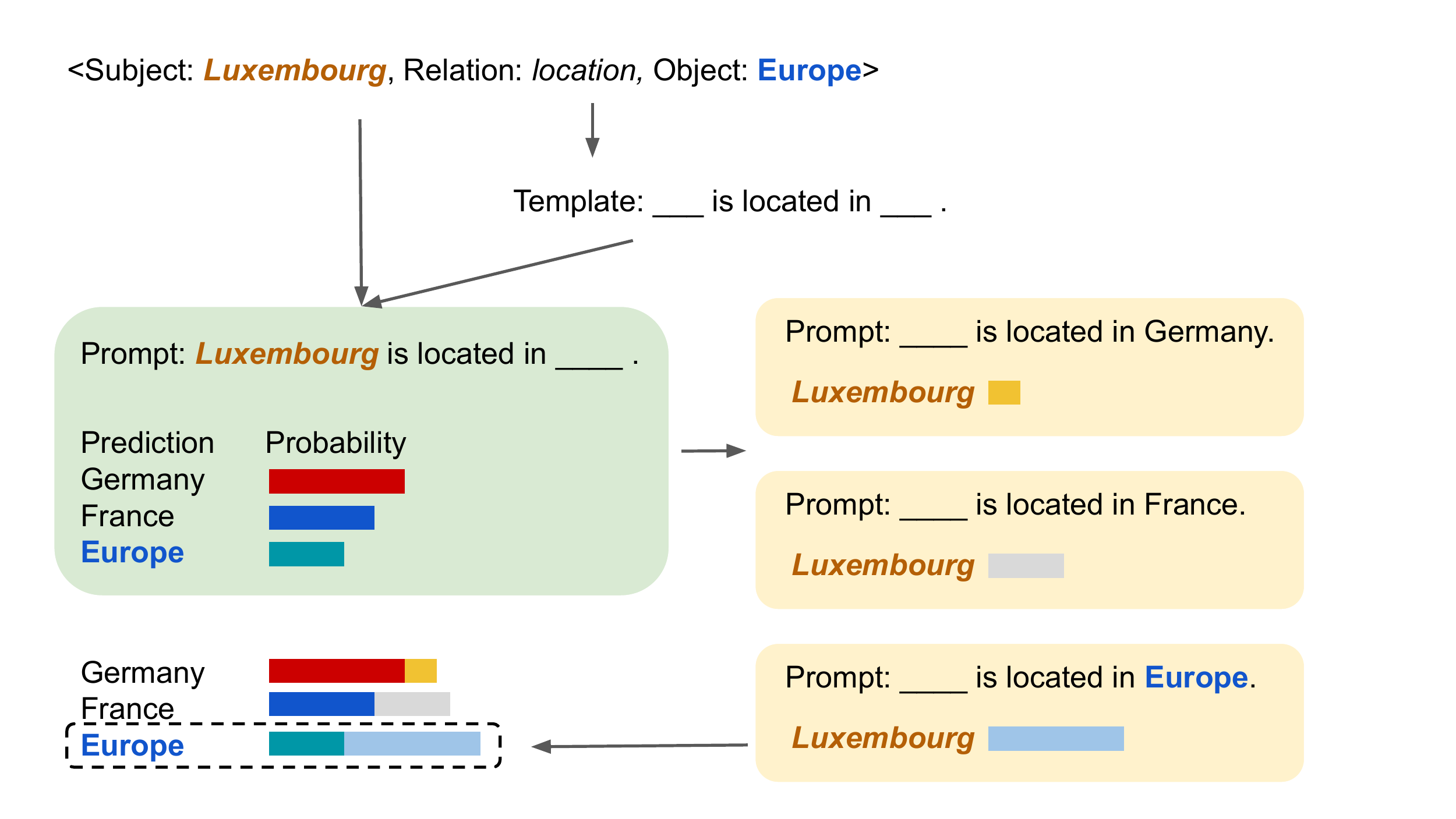}
	\caption{Example of factual probing: Given a subject and relation, predict the object. SPE uses a fixed template to generate a prompt for predicting object given subject (green box) as well as several symmetrical prompts for predicting the subject given object candidates (yellow boxes). The final prediction is obtained using the likelihoods of the object candidates and of the given subject as obtained using the symmetrical prompts. Bars represent probabilities from BERT. SPE is a continuous prompt method but we use natural language prompts and template here for illustration. }
	\label{fig:example}
\end{figure}

Existing methods for learning prompts do not leverage the symmetry inherent in the task's definition. For example, while \textit{Luxembourg} is located in \textit{Europe}, \textit{Europe} contains \textit{Luxembourg}. Similar ideas have been used for learning prompts for relation classification~\cite{Han2021PTRPT} and other NLP tasks~\cite{Crawford1996SymmetryBreakingPF, Kiddon2015SymmetryBasedSP, he-etal-2017-learning, tanchip-etal-2020-inferring}.

In this work, we propose \textit{Symmetrical Prompt Enhancement} (SPE)-- a continuous prompting method that \textit{learns} prompt that incorporates the above mentioned symmetry. Specifically, in addition to generating a prompt to predict the object given the subject, SPE also generates an additional symmetrical prompt to predict the subject given the object. Using the first prompt (see green box in Fig.~\ref{fig:example}), SPE obtains a few high-probability candidate objects like \textit{Germany}, \textit{France}, and \textit{Europe}. Thereafter, for each object candidate, it generates a symmetrical prompt (see yellow boxes), and obtains the likelihood of the subject, \textit{Luxembourg}. At the heart of SPE is a prompt generation model that is trained by maximizing the joint likelihood of both the candidates as well as the subject (given the candidates). Our experiments on the factual probing dataset LAMA \cite{petroni-etal-2019-language} show that SPE achieves significant improvement over previous approaches and our analysis points to sources of this performance gain. These experiments demonstrate that like SPE, probing methods should learn prompts that leverage the symmetry of the task because that can help PLMs in producing better answers when they are being probed for stored factual knowledge. 

\section{Symmetrical Prompt  Enhancement}
The goal of factual probing via prompt generation is to output object $\mathcal{O}$ for given subject  $\mathcal{I}$ and relation $\mathcal{R}$ by constructing a prompt $\mathcal{P}$. Most methods operate by assuming a template $\mathcal{T}$, and generating the prompt $\mathcal{P}$ from $\mathcal{T}$, $\mathcal{I}$ and $\mathcal{R}$. Fig.~\ref{fig:example} shows an example of Subject (\textit{Luxembourg}), Relation (\textit{location}), Object (\textit{Europe}), Template (\textit{\_\_\_\_ is located in \_\_\_\_.}), and the corresponding Prompt (\textit{Luxembourg is located in \_\_\_\_.}). The figure shows a natural language template and prompts for readability. However, for continuous prompt methods like ours, the template is a sequences of vectors like $[V]_{1} \dots [V]_{n}$ \_\_\_\_ $[V]_{n+1} \dots [V]_{n+m}$ \_\_\_\_ $[V]_{n+m+1} \dots [V]_{n+m+k}$, $\forall [V]_i \in \mathbb{R}^{d}$. We refer to the two blanks as $\text{B}_{\mathcal{O}}$ and $\text{B}_{\mathcal{I}}$. The prompt, $\mathcal{P}_{orig}$, is typically generated by learning these vectors and filling the (representation of) $\mathcal{I}$ in $\text{B}_{\mathcal{I}}$. The prompts are relation-specific ($\mathcal{P}_{orig}^\mathcal{R}$) but here we refer to them as $\mathcal{P}_{orig}$ for simplicity. The model's prediction, $\mathcal{\hat{O}}$, is the most likely object candidate for the $\text{B}_{\mathcal{O}}$ as determined by the PLM using $\mathcal{P}_{orig}$.

Our proposed approach, \textit{Symmetrical Prompt Enhancement} (SPE), leverages the inherent symmetry of the task. Specifically, in addition to learning the original prompt $\mathcal{P}_{orig}$ for predicting the object given the subject, SPE also generates several symmetrical prompts, $\mathcal{P}_{sym}$, for predicting the subject given the object. Like $\mathcal{P}_{orig}$, $\mathcal{P}_{sym}$ is also generated from $\mathcal{T}$ except that this time $\text{B}_{\mathcal{O}}$ is filled by the (representation of) $\mathcal{O}$. The prompt is used for probing the PLM which outputs prediction for $\text{B}_{\mathcal{I}}$.
\begin{gather}
    p(v|\mathcal{P}_{\orig}) = \prob_{\PLM}(\text{B}_{\mathcal{O}}=v|\mathcal{P}_{\orig}) \\
    p(v'|\mathcal{P}_{\sym}) = \prob_{\PLM}(\text{B}_{\mathcal{I}}=v'|\mathcal{P}_{\sym})
\end{gather}
Here $p(v|\mathcal{P})$ is the probability distribution of word or phrases $v$ in PLM given prompt $\mathcal{P}$ as input. The model is trained by optimizing a linear combination of the cross-entropy objectives of predicting the object $\mathcal{O}$ and the subject $\mathcal{I}$: 
\begin{equation}
\begin{aligned}
\max_{\theta}\log p(v=\mathcal{O}|\mathcal{P}_{orig}) + \lambda \log p(v'=\mathcal{I}|\mathcal{P}_{sym}),
\end{aligned}
\end{equation}
where $\lambda$ is a hyperparameter. $\theta$, the parameters of the prompt generation model, are learned. 

For inference, SPE selects top $K$ predictions $\mathcal{C}^{\textit{K}}$: 
\begin{align}
    \mathcal{C}^{\textit{K}} & = \TopK_{v \in \mathcal{V}} p(v|\mathcal{P}_{\orig}) \label{eq:candK}
\end{align}
and uses each prediction $c^{\textit{k}} \in \mathcal{C}^{\textit{K}}$ as a candidate to generate the symmetrical prompt $\mathcal{P}_{sym}^{\textit{k}}$. 
Finally, the model's prediction $\mathcal{\hat{O}}$ is: 
\begin{equation}
\begin{aligned}
    \mathcal{\hat{O}}  = & \arg\max_{c^{\textit{k}} \in \mathcal{C}^{\textit{K}}} \log p(v=c^{\textit{k}}|\mathcal{P}_{\orig})\\ 
    & + \lambda \log p({v'=\mathcal{I}}|\mathcal{P}_{\sym}^{\textit{k}}).
\end{aligned}
\end{equation}
In practice,  $\mathcal{L}$ and $\prob_{\PLM}$  are normalized by input length to account for inputs with multiple tokens. 
\begin{table*}[ht!]
\centering
\scalebox{0.85}{
\begin{tabular}{c|c|c|c|c|c|c|c|c|c}
\toprule
    \multirow{2}{*}{\textbf{Model}} &
    \multicolumn{3}{c|}{\textbf{BERT-base}} & 
    \multicolumn{3}{c|}{\textbf{BERT-large}} &
    \multicolumn{3}{c}{\textbf{RoBERTa-base}} \\
    \cmidrule{2-10} 
    & P@1 & P@10 & MRR & P@1 & P@10 & MRR & P@1 & P@10 & MRR\\
    \midrule
    Manual \cite{petroni-etal-2019-language}& 31.1 & 59.5 & 40.3 & 28.9 & 57.7 & 38.7 & 22.0 & 36.0 & 25.0\\
    \midrule
    LPAQA \cite{jiang-etal-2020-know}& 34.1 & 62.0 & 43.6 & 39.4 & 67.4 & 49.1 & 21.7 & 36.0 & 27.7\\
    \midrule
    AutoPrompt \cite{shin-etal-2020-autoprompt}& 43.3 & 73.9 & 53.9 & 41.3 & 69.3 & 50.6 & 40.0 & 68.3 & 49.9\\
    \midrule
    OptiPrompt (manual) \cite{zhong2021factual}& 48.6 & 79.0 & 58.9 & 50.6 & 79.2 & 60.7 & 40.3 & 65.7 & 48.9\\
    \midrule
    SoftPrompt (mined) \cite{qin-eisner-2021-learning} & 48.8 & 79.6 & 59.4 & 51.0 & 81.4 & 59.6 & 40.6 & 75.5 & 53.0 \\
    \midrule
    P-tuning \cite{Liu2021GPTUT}& 48.2 & 78.1 & 58.6 & 49.9 & 80.6 & 60.6 & 43.5 & 73.9 & 53.8\\
    \midrule
    SPE & \textbf{50.3} & \textbf{80.5} & \textbf{60.9} & \textbf{53.1} & \textbf{82.4} & \textbf{63.4} & \textbf{47.0} & \textbf{75.8} & \textbf{56.2}\\
    \bottomrule
\end{tabular}
}
\caption{SPE outperforms state-of-the-art discrete and continuous prompt approaches on the LAMA dataset.}
\label{tab:result} 
\end{table*}

\begin{table}
\centering
\scalebox{0.95}{
\begin{tabular}{c|c|c|c}
\toprule
    \textbf{Model} & P@1 & P@10 & MRR\\
    \midrule
    P-tuning & 48.2 & 78.1 & 58.6\\
    \midrule
    \textit{SPE}\hspace{1mm} K=1 & 48.7 & 79.9 & 59.5 \\
    \hspace{8mm} K=5 & 49.9 & 79.9 & 60.5 \\
    \hspace{10mm} K=10 & 49.9 & 79.9 & 60.7 \\
    \hspace{10mm} K=15 & \textbf{50.3} & \textbf{80.5} & \textbf{60.9} \\
    \bottomrule
\end{tabular}
}
\caption{Effect of varying size of candidate pool on SPE's performance. SPE outperforms P-tuning even without reranking (K=1). A larger candidate pool helps the model even further.}
\label{tab:reranking-bert-base}
\end{table}

\begin{table*}[!h]
\centering
\scalebox{0.8}{
\begin{tabular}{lccccccc}
\toprule
    \textbf{Relation} & \textbf{Subject} & \multicolumn{5}{l}{\textbf{Top 5 Predictions (Prob. High $\longrightarrow$ Low): Top - PT,   Bottom - SPE}} & \textbf{Rank}\\
    \midrule
    {P108} & \multirow{2}{*}{Spike Milligan} & Microsoft & IBM & Google & \underline{{BBC}} & ESPN & 4\\
    (employer) & & \underline{{BBC}} & Microsoft & CBS & ESPN & Google & 1\\
    \midrule
    {P364} & \multirow{2}{*}{Baaz} & Turkish & English & French & Arabic & Persian & 41\\
    (original language) & & \underline{{Hindi}} & Urdu & Punjabi & Bengali & Persian & 1\\
    \midrule
    {P101} & \multirow{2}{*}{Richard Wagner} & music & history & psychology & \underline{{opera}} & linguistics & 4\\
    (field of work) & & \underline{{opera}} & music & philosophy & aesthetics & art & 1\\
    \midrule
    {P27} & \multirow{2}{*}{Rubens Barrichello} & Belgium & France & Italy & Spain & Germany & 15\\
    (country of citizenship) & & \underline{{Brazil}} & Spain & Argentina & Portugal & Uruguay & 1\\
    \midrule
    {P30} & \multirow{2}{*}{Marshall Islands} & Antarctica & Asia & Africa & \underline{{Oceania}} & Europe & 4\\
    (continent) & & Asia & \underline{{Oceania}} & Africa & Antarctica & Europe & 2\\
    \midrule
    {P279} & \multirow{2}{*}{river} & \underline{{river}} & stream & tributary & canal & creek & 1\\
    (subclass of) & & tributary & stream & \underline{{river}} & creek & tributaries & 3\\
    \bottomrule
\end{tabular}
}
\caption{Sample outputs of P-tuning (PT) and SPE. The ranks of correct answers (\underline{{underlined}}) are in the last column.}
\label{tab:example}
\end{table*}

\section{Implementation Details}
\label{sec: experiment setup}
We conduct experiments on the fact retrieval part of LAMA dataset~\cite{petroni-etal-2019-language}, which consists of fact triples with single-token objects from 41 relations in Wikidata \cite{10.1145/2629489}. We use the  training set extended by \citet{shin-etal-2020-autoprompt}. We choose masked language models BERT \cite{devlin-etal-2019-bert} and RoBERTa \cite{Liu2019RoBERTaAR} as PLMs, which are fixed during training to serve as static knowledge bases. For implementation, we use PLMs in Huggingface library of Transformers \cite{wolf-etal-2020-transformers}. We follow \citet{Liu2021GPTUT} for designing templates and the prompt generation component of our model. In particular, we use BiLSTM~\cite{Graves2013SpeechRW} with multilayer perceptron (MLP) for prompt generation and use the following generic and relation-agnostic format for template, $\mathcal{T}$: \textit{$[V]_{1}$ $[V]_{2}$ $[V]_{3}$ \_\_\_\_ $[V]_{4}$ $[V]_{5}$ $[V]_{6}$ \_\_\_\_ $[V]_{7}$ $[V]_{8}$ $[V]_{9}$ $\forall [V]_i \in \mathbb{R}^{d}$}. The model and the template are randomly initialized.

For $\mathcal{I}$ with multiple tokens, we mask them one token at a time to generate $\mathcal{P}_{\sym}$, and use the average of pseudo likelihoods from all $\mathcal{P}_{\sym}$s to represent $\log p(v'=\mathcal{I}|\mathcal{P}_{\sym})$. In practice, we find that masking one token at a time is better than masking the entire phrase at once, and averaging the pseudo-likelihood has better performance. The training batch size is 8. We set K to be 15 during inference, and $\lambda$ to be 0.8 based on our experiments on the development set. 
The results are evaluated by accuracy at top 1 (P@1) and top 10 (P@10) predictions, and Mean Reciprocal Rank (MRR) as in~\citet{qin-eisner-2021-learning}. Appendix \ref{sec: implementation} includes more setup details and discussion on choice of $\lambda$.

\section{Results}
\label{sec: result}
We compare our results with both discrete and continuous prompt methods. Discrete prompt methods include prompts from manually designed templates \cite{petroni-etal-2019-language}; LPAQA \cite{jiang-etal-2020-know}, which uses text mining based prompts; and AutoPrompt~\cite{shin-etal-2020-autoprompt}, which uses discrete lexicalized trigger tokens for prompt generation. Continuous prompt methods include P-tuning~\cite{Liu2021GPTUT}, which uses a neural network to generate prompts; OptiPrompt~\cite{zhong2021factual}, which uses manually initialized prompts; and SoftPrompt~\cite{qin-eisner-2021-learning}, which ensembles multiple prompts initialized with mined templates.

\noindent \textbf{Quantitative Results: }Table \ref{tab:result} shows the performance of SPE and all baselines. The results show that SPE outperforms all previous methods. Note that, unlike OptiPrompt and SoftPrompt, SPE does not make use of manually designed templates for initialization. We also find that SPE outperforms the baselines when the PLM parameters are updated jointly with the prompt tokens on the training data. See Table~\ref{tab: LAMA-finetune} in the Appendix~\ref{appendix: finetuning} for detailed results. For the rest experiments, we consider P-tuning as primary baseline since it is the best performing model that is directly comparable to SPE. 

\noindent \textbf{Effect of candidates pool size:} Table~\ref{tab:reranking-bert-base} shows how SPE performs with different candidate pool sizes. Comparing the first two rows we can see that SPE outperforms our primary baseline, P-tuning, even without reranking (K=1). Increasing the size of the candidate pool leads to further improvements. However, expanding the candidate pool has a trade-off between performance and memory usage. Meanwhile, applying reranking on the discrete prompt methods mentioned does not introduce performance gain, mainly because their prompt templates are selected or mined in favor of object prediction only. We leave the investigation of constructing discrete prompts that benefits from the symmetry as future work. 

\noindent \textbf{Performance on Easy and Hard examples: }
The LAMA test set has also been split into LAMA-Easy and LAMA-Hard where objects in the LAMA-Easy split can be "guessed" by naive methods~\cite{zhong2021factual}.
We observe that SPE outperforms the baselines in P@1 for both splits and its gain over P-tuning for LAMA-Hard (4.2\%) is larger than LAMA-Easy (1.5\%) (see Table~\ref{tab: LAMA-Easy-Hard} of Appendix~\ref{appendix: LAMA-Easy LAMA-Hard}).  This indicates that the improvement of SPE does not simply come from shallow pattern matching and it performs well on hard examples.

\begin{figure}[t!]
	\centering
	\includegraphics[width=0.5
	\textwidth]{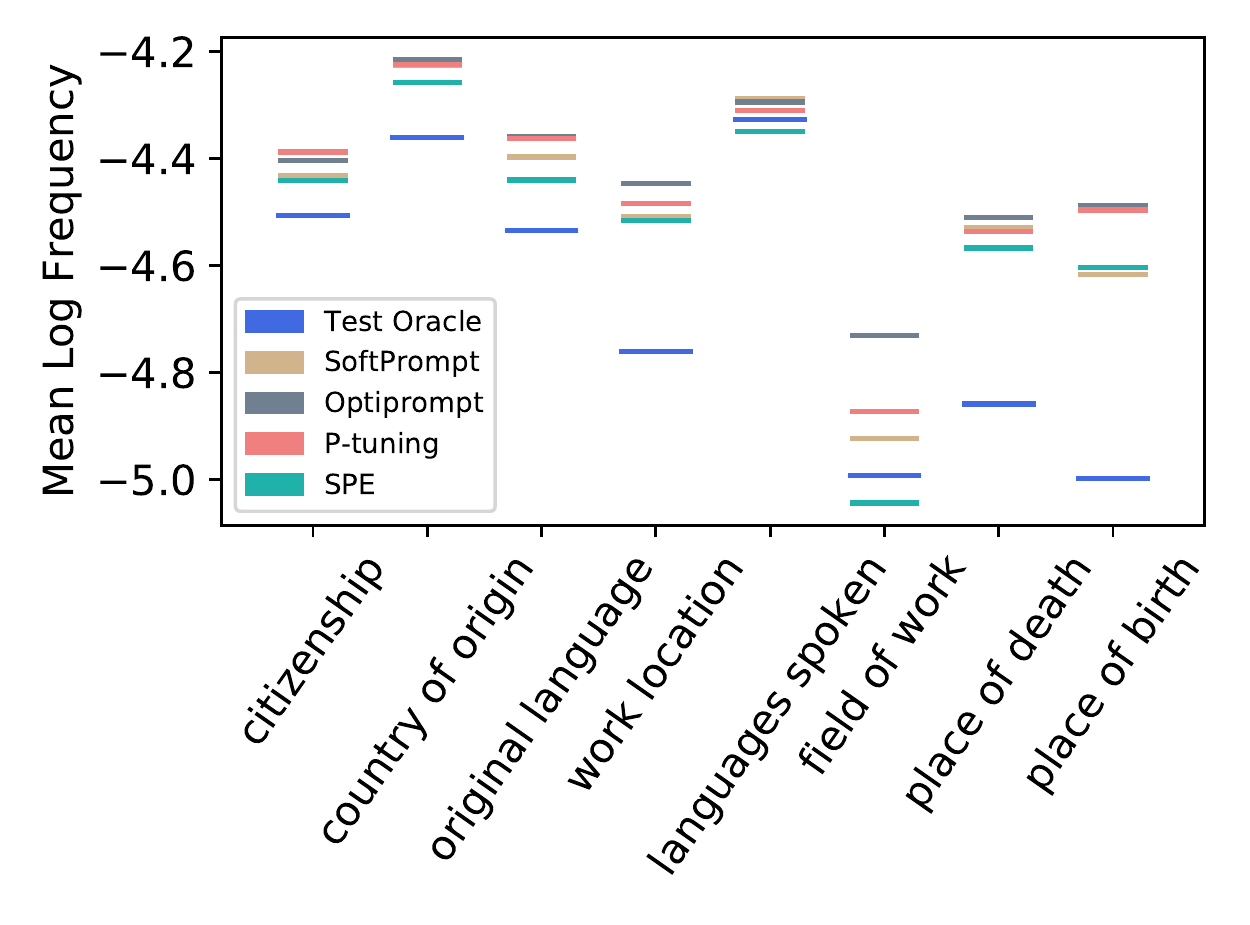}
	\caption{Frequencies of predictions of different methods. SPE can output answers that have low frequencies.}
	\label{fig:mean freq top1}
\end{figure}

\noindent \textbf{Qualitative Results and Analysis: } We include some qualitative examples of top 5 predictions in Table~\ref{tab:example} from P-tuning (top half of each row) and SPE (bottom half of each row). The correct answers are underlined, and their ranks in the predicted lists are in the last column. We observe that SPE's top predictions are in the correct domain. For example, SPE outputs \textit{BBC} for the \textit{employer} of British-Irish actor \textit{Spike Milligan} (as opposed to \textit{Microsoft}, \textit{IBM}, and \textit{Google}) as outputted by P-tuning), and \textit{Hindi} along with other Indian languages, when asked about the \textit{original language} of an Indian movie \textit{Baaz}, rather than \textit{Turkish}, a non-Indian Language. Moreover, SPE correctly identifies the \textit{country of citizenship} for \textit{Rubens Barrichello} as \textit{Brazil}. Identifying objects for relations like \textit{country of citizenship} for individuals are challenging because documents with the individual's names in the pretraining corpus of PLMs might contain mentions of multiple places he/she has worked or lived or received education in. Therefore, these co-occurrences might confuse PLMs. In Appendix~\ref{sec: preciseness}, we identify such \textit{confusing} relations and conduct a close analysis on them.

We also find that SPE's predictions (e.g. \textit{opera} for the field of work of \textit{Richard Wagner}) are more precise that P-tuning's (\textit{music}). In general, PLMs predictions for a relation can get affected by related high-frequency but incorrect object candidates. Previous prompt methods are found to suffer from bias of the prompt and object distribution in the dataset~\cite{cao-etal-2021-knowledgeable}. To investigate this, we identify a set of relations that are prone to such spurious frequency-related associations (see Appendix~\ref{sec: preciseness} for the list of relations) and find that SPE especially performs well on such relations (see Figure~\ref{fig:rels_highlighted} and Appendix~\ref{sec:preciseness_comparison}). We also plot the mean ($\log_{10}$) token frequencies of top predictions of different methods as well as the oracle for these relations in Figure ~\ref{fig:mean freq top1} (using word frequencies from \citet{robyn_speer_2018_1443582}). We observe that SPE's predictions (green bars) have lower frequencies than most baselines including P-tuning (red bars). Meanwhile, the frequencies of SPE's predictions are in general more similar to that of the oracle (blue bars)  than most baselines. This indicates that even though the correct (and more precise) answers have lower frequencies, SPE can output them as answers while the baselines output the more frequent alternatives as answers (see Appendix~\ref{sec:preciseness_top1} for examples). We further extend this analysis to the top M predictions and observe similar behavior (see Appendix~\ref{sec:preciseness_topM}). Lastly, the outputs of SPE are less affected by the most frequently occurring objects in the dataset (see Appendix~\ref{sec:label_dist_association}).

\section{Limitations}
We note that SPE may not help if the correct objects are broad concepts (e.g. "mathematics" vs "algebra", "river" vs "tributary", "FIFA" vs "UEFA"). Typical relations with such objects include P279 \textit{subclass of}, P361 \textit{part of} and P463 \textit{member of}. The top 5 predictions by SPE (and also P-tuning) for the \textit{subclass of} relation are shown in Table~\ref{tab:example}. The correct answer, \textit{river}, is ranked 3rd by SPE and an incorrect answer, \textit{tributary}, is the top prediction. P-tuning outputs the correct answer. 

Also, in general, SPE can get affected by error propagation because of its two-step inference process that  first predicts object candidates and then ranks them.

Though the proposed symmetrical prompt method improves knowledge probing, the utility of the technique in other NLP tasks is not yet investigated. Besides, the experiments are only conducted for masked language models but there has been recent progress in other types of language models which are not explored in the paper. Lastly, the proposed method requires additional computational cost compared the baselines.

\section{Conclusion}
This work introduces Symmetrical Prompt Enhancement (SPE) -- a continuous prompt-learning method for factual probing of PLMs by learning prompts that utilize the inherent symmetry of the task. Our experiments show that SPE outperforms existing SOTA methods thereby helping us know more about how much knowledge is stored in a PLM. Future work could explore this idea of using task symmetry for other  NLP tasks. \footnote{Code is available \href{https://github.com/Nativeatom/SPE}{here}.}

\section{Ethical Consideration}
In this work, we propose SPE, which incorporates the symmetrical nature of factual knowledge in prompt methods. Our result shows the effectiveness of SPE over several previous prompt baselines. Even though we work on the factual knowledge dataset, we notice that current PLMs does not have the awareness to distinguish between publicly-available factual knowledge and private information (which is not considered as knowledge) either during the pre-training or inference, while the memorizing information of PLMs in latter lead to potential risk of privacy leakage \cite{carlini21extracting}. All the experiments are conducted on the publicly available dataset, which is mainly based on Wikidata.

\bibliography{anthology,custom}

\begin{thebibliography}{29}
\expandafter\ifx\csname natexlab\endcsname\relax\def\natexlab#1{#1}\fi

\bibitem[{Bouraoui et~al.(2020)Bouraoui, Camacho-Collados, and
  Schockaert}]{Bouraoui2020InducingRK}
Zied Bouraoui, Jose Camacho-Collados, and Steven Schockaert. 2020.
\newblock \href {https://doi.org/10.1609/aaai.v34i05.6242} {Inducing relational
  knowledge from bert}.
\newblock \emph{Proceedings of the AAAI Conference on Artificial Intelligence},
  34(05):7456--7463.

\bibitem[{Cao et~al.(2021)Cao, Lin, Han, Sun, Yan, Liao, Xue, and
  Xu}]{cao-etal-2021-knowledgeable}
Boxi Cao, Hongyu Lin, Xianpei Han, Le~Sun, Lingyong Yan, Meng Liao, Tong Xue,
  and Jin Xu. 2021.
\newblock \href {https://doi.org/10.18653/v1/2021.acl-long.146} {Knowledgeable
  or educated guess? revisiting language models as knowledge bases}.
\newblock In \emph{Proceedings of the 59th Annual Meeting of the Association
  for Computational Linguistics and the 11th International Joint Conference on
  Natural Language Processing (Volume 1: Long Papers)}, pages 1860--1874,
  Online. Association for Computational Linguistics.

\bibitem[{Carlini et~al.(2021)Carlini, Tramer, Wallace, Jagielski,
  Herbert-Voss, Lee, Roberts, Brown, Song, Erlingsson, Oprea, and
  Raffel}]{carlini21extracting}
Nicholas Carlini, Florian Tramer, Eric Wallace, Matthew Jagielski, Ariel
  Herbert-Voss, Katherine Lee, Adam Roberts, Tom Brown, Dawn Song, Ulfar
  Erlingsson, Alina Oprea, and Colin Raffel. 2021.
\newblock \href {https://arxiv.org/abs/2012.07805} {Extracting training data
  from large language models}.
\newblock In \emph{USENIX Security Symposium}.

\bibitem[{Chang and Bergen(2022)}]{Chang2021WordAI}
Tyler~A. Chang and Benjamin~K. Bergen. 2022.
\newblock \href {https://doi.org/10.1162/tacl_a_00444} {Word acquisition in
  neural language models}.
\newblock \emph{Transactions of the Association for Computational Linguistics},
  10:1--16.

\bibitem[{Crawford et~al.(1996)Crawford, Ginsberg, Luks, and
  Roy}]{Crawford1996SymmetryBreakingPF}
James~M. Crawford, Matthew~L. Ginsberg, Eugene~M. Luks, and Amitabha Roy. 1996.
\newblock Symmetry-breaking predicates for search problems.
\newblock In \emph{Proceedings of the Fifth International Conference on
  Principles of Knowledge Representation and Reasoning}, KR'96, page 148–159,
  San Francisco, CA, USA. Morgan Kaufmann Publishers Inc.

\bibitem[{Devlin et~al.(2019)Devlin, Chang, Lee, and
  Toutanova}]{devlin-etal-2019-bert}
Jacob Devlin, Ming-Wei Chang, Kenton Lee, and Kristina Toutanova. 2019.
\newblock \href {https://doi.org/10.18653/v1/N19-1423} {{BERT}: Pre-training of
  deep bidirectional transformers for language understanding}.
\newblock In \emph{Proceedings of the 2019 Conference of the North {A}merican
  Chapter of the Association for Computational Linguistics: Human Language
  Technologies, Volume 1 (Long and Short Papers)}, pages 4171--4186,
  Minneapolis, Minnesota. Association for Computational Linguistics.

\bibitem[{Glorot et~al.(2011)Glorot, Bordes, and Bengio}]{Glorot2011DeepSR}
Xavier Glorot, Antoine Bordes, and Yoshua Bengio. 2011.
\newblock \href {https://proceedings.mlr.press/v15/glorot11a.html} {Deep sparse
  rectifier neural networks}.
\newblock In \emph{Proceedings of the Fourteenth International Conference on
  Artificial Intelligence and Statistics}, volume~15 of \emph{Proceedings of
  Machine Learning Research}, pages 315--323, Fort Lauderdale, FL, USA. PMLR.

\bibitem[{Graves et~al.(2013)Graves, rahman Mohamed, and
  Hinton}]{Graves2013SpeechRW}
Alex Graves, Abdel rahman Mohamed, and Geoffrey~E. Hinton. 2013.
\newblock Speech recognition with deep recurrent neural networks.
\newblock \emph{2013 IEEE International Conference on Acoustics, Speech and
  Signal Processing}, pages 6645--6649.

\bibitem[{Han et~al.(2021)Han, Zhao, Ding, Liu, and Sun}]{Han2021PTRPT}
Xu~Han, Weilin Zhao, Ning Ding, Zhiyuan Liu, and Maosong Sun. 2021.
\newblock Ptr: Prompt tuning with rules for text classification.
\newblock \emph{ArXiv}, abs/2105.11259.

\bibitem[{Haviv et~al.(2021)Haviv, Berant, and
  Globerson}]{haviv-etal-2021-bertese}
Adi Haviv, Jonathan Berant, and Amir Globerson. 2021.
\newblock \href {https://aclanthology.org/2021.eacl-main.316} {{BERT}ese:
  Learning to speak to {BERT}}.
\newblock In \emph{Proceedings of the 16th Conference of the European Chapter
  of the Association for Computational Linguistics: Main Volume}, pages
  3618--3623, Online. Association for Computational Linguistics.

\bibitem[{He et~al.(2017)He, Balakrishnan, Eric, and
  Liang}]{he-etal-2017-learning}
He~He, Anusha Balakrishnan, Mihail Eric, and Percy Liang. 2017.
\newblock \href {https://doi.org/10.18653/v1/P17-1162} {Learning symmetric
  collaborative dialogue agents with dynamic knowledge graph embeddings}.
\newblock In \emph{Proceedings of the 55th Annual Meeting of the Association
  for Computational Linguistics (Volume 1: Long Papers)}, pages 1766--1776,
  Vancouver, Canada. Association for Computational Linguistics.

\bibitem[{Jiang et~al.(2021)Jiang, Araki, Ding, and
  Neubig}]{10.1162/tacl_a_00407}
Zhengbao Jiang, Jun Araki, Haibo Ding, and Graham Neubig. 2021.
\newblock \href {https://doi.org/10.1162/tacl_a_00407} {{How Can We Know When
  Language Models Know? On the Calibration of Language Models for Question
  Answering}}.
\newblock \emph{Transactions of the Association for Computational Linguistics},
  9:962--977.

\bibitem[{Jiang et~al.(2020)Jiang, Xu, Araki, and
  Neubig}]{jiang-etal-2020-know}
Zhengbao Jiang, Frank~F. Xu, Jun Araki, and Graham Neubig. 2020.
\newblock \href {https://doi.org/10.1162/tacl_a_00324} {How can we know what
  language models know?}
\newblock \emph{Transactions of the Association for Computational Linguistics},
  8:423--438.

\bibitem[{Kiddon and Domingos(2015)}]{Kiddon2015SymmetryBasedSP}
Chlo{\'e} Kiddon and Pedro~M. Domingos. 2015.
\newblock Symmetry-based semantic parsing.

\bibitem[{Lester et~al.(2021)Lester, Al-Rfou, and
  Constant}]{lester-etal-2021-power}
Brian Lester, Rami Al-Rfou, and Noah Constant. 2021.
\newblock \href {https://doi.org/10.18653/v1/2021.emnlp-main.243} {The power of
  scale for parameter-efficient prompt tuning}.
\newblock In \emph{Proceedings of the 2021 Conference on Empirical Methods in
  Natural Language Processing}, pages 3045--3059, Online and Punta Cana,
  Dominican Republic. Association for Computational Linguistics.

\bibitem[{Liu et~al.(2022)Liu, Yuan, Fu, Jiang, Hayashi, and
  Neubig}]{Liu2021PretrainPA}
Pengfei Liu, Weizhe Yuan, Jinlan Fu, Zhengbao Jiang, Hiroaki Hayashi, and
  Graham Neubig. 2022.
\newblock \href {https://doi.org/10.1145/3560815} {Pre-train, prompt, and
  predict: A systematic survey of prompting methods in natural language
  processing}.
\newblock \emph{ACM Comput. Surv.}
\newblock Just Accepted.

\bibitem[{Liu et~al.(2021)Liu, Zheng, Du, Ding, Qian, Yang, and
  Tang}]{Liu2021GPTUT}
Xiao Liu, Yanan Zheng, Zhengxiao Du, Ming Ding, Yujie Qian, Zhilin Yang, and
  Jie Tang. 2021.
\newblock Gpt understands, too.
\newblock \emph{ArXiv}, abs/2103.10385.

\bibitem[{Liu et~al.(2019)Liu, Ott, Goyal, Du, Joshi, Chen, Levy, Lewis,
  Zettlemoyer, and Stoyanov}]{Liu2019RoBERTaAR}
Yinhan Liu, Myle Ott, Naman Goyal, Jingfei Du, Mandar Joshi, Danqi Chen, Omer
  Levy, Mike Lewis, Luke Zettlemoyer, and Veselin Stoyanov. 2019.
\newblock Roberta: A robustly optimized bert pretraining approach.
\newblock \emph{ArXiv}, abs/1907.11692.

\bibitem[{Newman et~al.(2022)Newman, Choubey, and Rajani}]{newman2022padapters}
Benjamin Newman, Prafulla~Kumar Choubey, and Nazneen Rajani. 2022.
\newblock \href {https://openreview.net/forum?id=DhzIU48OcZh} {P-adapters:
  Robustly extracting factual information from language models with diverse
  prompts}.
\newblock In \emph{International Conference on Learning Representations}.

\bibitem[{Ott et~al.(2018)Ott, Auli, Grangier, and Ranzato}]{pmlr-v80-ott18a}
Myle Ott, Michael Auli, David Grangier, and Marc'Aurelio Ranzato. 2018.
\newblock \href {https://proceedings.mlr.press/v80/ott18a.html} {Analyzing
  uncertainty in neural machine translation}.
\newblock In \emph{Proceedings of the 35th International Conference on Machine
  Learning}, volume~80 of \emph{Proceedings of Machine Learning Research},
  pages 3956--3965. PMLR.

\bibitem[{Petroni et~al.(2019)Petroni, Rockt{\"a}schel, Riedel, Lewis, Bakhtin,
  Wu, and Miller}]{petroni-etal-2019-language}
Fabio Petroni, Tim Rockt{\"a}schel, Sebastian Riedel, Patrick Lewis, Anton
  Bakhtin, Yuxiang Wu, and Alexander Miller. 2019.
\newblock \href {https://doi.org/10.18653/v1/D19-1250} {Language models as
  knowledge bases?}
\newblock In \emph{Proceedings of the 2019 Conference on Empirical Methods in
  Natural Language Processing and the 9th International Joint Conference on
  Natural Language Processing (EMNLP-IJCNLP)}, pages 2463--2473, Hong Kong,
  China. Association for Computational Linguistics.

\bibitem[{Qin and Eisner(2021)}]{qin-eisner-2021-learning}
Guanghui Qin and Jason Eisner. 2021.
\newblock \href {https://doi.org/10.18653/v1/2021.naacl-main.410} {Learning how
  to ask: Querying {LM}s with mixtures of soft prompts}.
\newblock In \emph{Proceedings of the 2021 Conference of the North American
  Chapter of the Association for Computational Linguistics: Human Language
  Technologies}, pages 5203--5212, Online. Association for Computational
  Linguistics.

\bibitem[{Schick and Sch{\"u}tze(2021)}]{schick-schutze-2021-exploiting}
Timo Schick and Hinrich Sch{\"u}tze. 2021.
\newblock \href {https://aclanthology.org/2021.eacl-main.20} {Exploiting
  cloze-questions for few-shot text classification and natural language
  inference}.
\newblock In \emph{Proceedings of the 16th Conference of the European Chapter
  of the Association for Computational Linguistics: Main Volume}, pages
  255--269, Online. Association for Computational Linguistics.

\bibitem[{Shin et~al.(2020)Shin, Razeghi, Logan~IV, Wallace, and
  Singh}]{shin-etal-2020-autoprompt}
Taylor Shin, Yasaman Razeghi, Robert~L. Logan~IV, Eric Wallace, and Sameer
  Singh. 2020.
\newblock \href {https://doi.org/10.18653/v1/2020.emnlp-main.346}
  {{A}uto{P}rompt: {E}liciting {K}nowledge from {L}anguage {M}odels with
  {A}utomatically {G}enerated {P}rompts}.
\newblock In \emph{Proceedings of the 2020 Conference on Empirical Methods in
  Natural Language Processing (EMNLP)}, pages 4222--4235, Online. Association
  for Computational Linguistics.

\bibitem[{Speer et~al.(2018)Speer, Chin, Lin, Jewett, and
  Nathan}]{robyn_speer_2018_1443582}
Robyn Speer, Joshua Chin, Andrew Lin, Sara Jewett, and Lance Nathan. 2018.
\newblock \href {https://doi.org/10.5281/zenodo.1443582}
  {Luminosoinsight/wordfreq: v2.2}.

\bibitem[{Tanchip et~al.(2020)Tanchip, Yu, Xu, and
  Xu}]{tanchip-etal-2020-inferring}
Chelsea Tanchip, Lei Yu, Aotao Xu, and Yang Xu. 2020.
\newblock \href {https://doi.org/10.18653/v1/2020.findings-emnlp.259}
  {Inferring symmetry in natural language}.
\newblock In \emph{Findings of the Association for Computational Linguistics:
  EMNLP 2020}, pages 2877--2886, Online. Association for Computational
  Linguistics.

\bibitem[{Vrande\v{c}i\'{c} and Kr\"{o}tzsch(2014)}]{10.1145/2629489}
Denny Vrande\v{c}i\'{c} and Markus Kr\"{o}tzsch. 2014.
\newblock \href {https://doi.org/10.1145/2629489} {Wikidata: A free
  collaborative knowledgebase}.
\newblock \emph{Commun. ACM}, 57(10):78–85.

\bibitem[{Wolf et~al.(2020)Wolf, Debut, Sanh, Chaumond, Delangue, Moi, Cistac,
  Rault, Louf, Funtowicz, Davison, Shleifer, von Platen, Ma, Jernite, Plu, Xu,
  Le~Scao, Gugger, Drame, Lhoest, and Rush}]{wolf-etal-2020-transformers}
Thomas Wolf, Lysandre Debut, Victor Sanh, Julien Chaumond, Clement Delangue,
  Anthony Moi, Pierric Cistac, Tim Rault, Remi Louf, Morgan Funtowicz, Joe
  Davison, Sam Shleifer, Patrick von Platen, Clara Ma, Yacine Jernite, Julien
  Plu, Canwen Xu, Teven Le~Scao, Sylvain Gugger, Mariama Drame, Quentin Lhoest,
  and Alexander Rush. 2020.
\newblock \href {https://doi.org/10.18653/v1/2020.emnlp-demos.6} {Transformers:
  State-of-the-art natural language processing}.
\newblock In \emph{Proceedings of the 2020 Conference on Empirical Methods in
  Natural Language Processing: System Demonstrations}, pages 38--45, Online.
  Association for Computational Linguistics.

\bibitem[{Zhong et~al.(2021)Zhong, Friedman, and Chen}]{zhong2021factual}
Zexuan Zhong, Dan Friedman, and Danqi Chen. 2021.
\newblock \href {https://doi.org/10.18653/v1/2021.naacl-main.398} {Factual
  probing is [{MASK}]: Learning vs. learning to recall}.
\newblock In \emph{Proceedings of the 2021 Conference of the North American
  Chapter of the Association for Computational Linguistics: Human Language
  Technologies}, pages 5017--5033, Online. Association for Computational
  Linguistics.

\end{thebibliography}
\bibliographystyle{acl_natbib}

\newpage
\appendix

\begin{figure*}[t!]
	\centering
	\includegraphics[width=0.85\textwidth]{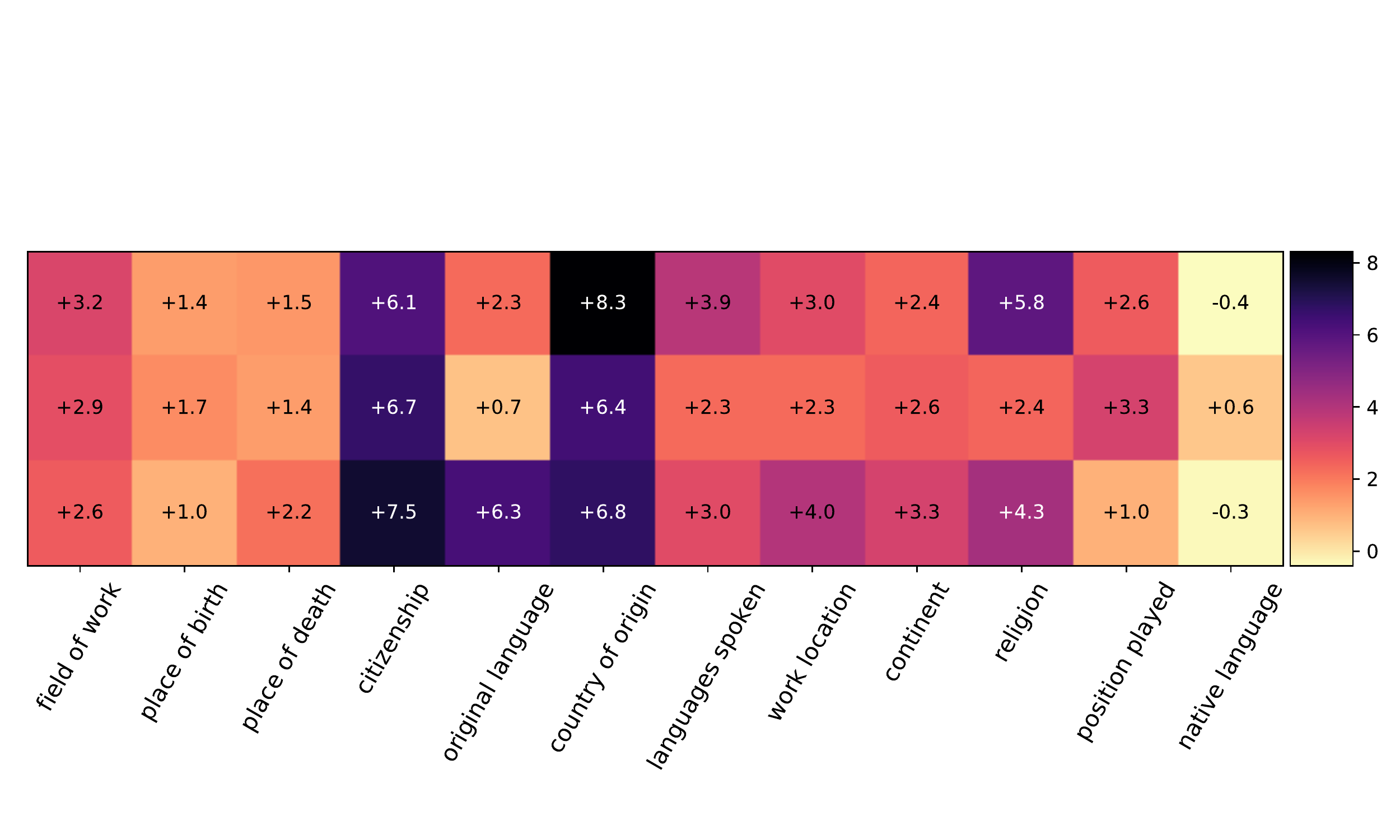}
	\caption{P@1 improvement of SPE under different relations in BERT-large-cased (scale of 100, darker color represents higher value): P101 \textit{field of work} (N-M), P19 \textit{place of birth} (N-1), P20 \textit{place of death}, P27 \textit{citizenship} (N-M), P364 \textit{original language of film or TV show} (N-1), P495 \textit{country of origin} (N-1), P1412 \textit{language spoken} (N-M), P937 \textit{work location} (N-M), P30 \textit{continent} (N-1), P140 \textit{religion} (N-1), P413 \textit{position played} (N-1), and P103 \textit{native language} (N-1). The first, second and third row represents the P@1 improvement SPE has over Optiprompt, SoftPrompt and P-tuning respectively. SPE outperforms all three probing methods for most relations.}
	\label{fig:rels_highlighted}
\end{figure*}

\section{Additional Implementation Details}
\label{sec: implementation}
\noindent \textbf{Prompt Generation Model:} The prompt generation model is based on work by \citet{Liu2021GPTUT}. It consists of a two-layer BiLSTM and a two-layer MLP on top of it. The MLP uses ReLU ~\cite{Glorot2011DeepSR} as the activation function. The hidden size of LSTM and dimension of $d$ are 768 for BERT-base-cased and RoBERTa-base, and 1024 for BERT-large-cased. The max training epoch is 100, and training stops when development performance does not increase for 20 epochs. The optimizer is Adam with learning rate being 1e-5. Other setting also follows \citet{Liu2021GPTUT}. The number of parameters is determined by the PLMs: BERT-base-cased (110M), BERT-large-cased (340M) and RoBERTa-base (125M); and the prompt generation model (14M). The experiments require 20 hours to finish on a single Tesla V100 GPU.

Also, during our experiments, we experiment with having separate prompt generation models for generating $\mathcal{P}_{orig}$ and $\mathcal{P}_{sym}$. However, we find that training one prompt generation model for both $\mathcal{P}_{orig}$ and $\mathcal{P}_{sym}$ led to better results.

\noindent \textbf{Choice of $\lambda$:} In our preliminary experiments on the development set, we find $\lambda=0.8$ to be the best choice among $[0, 1]$. However, we observe that the performance is not very sensitive to $\lambda$ and $\lambda > 0.4$ generally gives a reasonable performance. 

\section{Additional Results}

\begin{table*}
\centering
\scalebox{0.85}{
\begin{tabular}{c|c|c|c|c|c|c}
\toprule
    \textbf{Dataset} & \multicolumn{3}{c|}{\textbf{LAMA-Easy}} &  \multicolumn{3}{c}{\textbf{LAMA-Hard}}\\
    \midrule
    \textbf{Model} & P@1 & P@10 & MRR & P@1 & P@10 & MRR\\
    \midrule
    Manual & 40.2 & 70.7 & 47.9 & 27.1 & 54.2 & 35.1\\
    \midrule
    LPAQA & 46.0 & 71.7 & 52.6 & 27.3 & 55.7 & 35.3\\
    \midrule
    AutoPrompt & 58.2 & 85.7 & 66.2 & 28.6 & 60.9 & 39.5\\
    \midrule
    Optiprompt & 73.2 & 94.8 & 81.3 & 37.6 & 73.6 & 50.4\\
    \midrule
    SoftPrompt & 77.0 & \textbf{96.0} & \textbf{83.9} & 38.4 & 76.8 & 51.7\\
    \midrule
    P-tuning & 75.9 & 94.1 & 82.9 & 35.2 & 75.3 & 49.2\\
    \midrule
    SPE & \textbf{77.4} & 94.4 & 83.6 & \textbf{39.4} & \textbf{77.5} & \textbf{52.9} \\
    \bottomrule
\end{tabular}
}
\caption{SPE outperforms the baselines on both hard and easy examples with greater improvement on the hard ones.}
\label{tab: LAMA-Easy-Hard}
\end{table*}

\subsection{Easy and Hard LAMA Examples}
\label{appendix: LAMA-Easy LAMA-Hard}
\citet{zhong2021factual} points out that during factual probing, a PLM's predictions can be based on shallow patterns in the training data instead of the knowledge stored in the PLM. To study this phenomena, they propose an \textit{easy} (LAMA-Easy) and a \textit{hard} (LAMA-Hard) split of the LAMA dataset where objects in the LAMA-Easy subset can be "guessed" by naive or non-pretrained models. We compare SPE with the baselines on these two subsets and report results in Table~\ref{tab: LAMA-Easy-Hard}. We observe that, in general, all methods achieves better performance in LAMA-Easy that the complete testset but SPE has the highest P@1. It is outperformed on P@10 and MRR only by Softprompt and Optiprompt but they use manually designed templates. More importantly, SPE shows higher improvement in LAMA-Hard compared to baselines especially with respect to P-tuning (4.2\%  in P@1). This shows that the improvement of SPE does not simply come from shallow pattern matching and it is better at handling more challenging knowledge probing cases. 

\begin{table}
\centering
\scalebox{0.85}{
\begin{tabular}{c|c|c|c}
\toprule
    \textbf{Model} & P@1 & P@10 & MRR\\
    \midrule
    AutoPrompt & 41.3 & 61.6 & 50.6\\
    \midrule
    Optiprompt & 53.3 & 74.9 & 63.3\\
    \midrule
    SoftPrompt & 51.6 & 81.9 & 62.1\\
    \midrule
    P-tuning & 51.4 & 82.1 & 61.8\\
    \midrule
    SPE & \textbf{53.7} & \textbf{83.0} & \textbf{63.9} \\
    \bottomrule
\end{tabular}
}
\caption{Performance comparison of gradient-based prompt methods when the PLM is finetuned. SPE outperforms the baselines.}
\label{tab: LAMA-finetune}
\end{table}

\subsection{Finetuning PLMs}
\label{appendix: finetuning}
In the experiments reported in the paper, the PLMs are fixed during training and only the prompt generation model is being trained. We now experiment with also finetuning the PLMs. We use BERT-large-cased for this experiment. The results are reported in Table~\ref{tab: LAMA-finetune} where we compare SPE with the comparable gradient-based baselines. We can see that SPE outperforms those baselines in this setting also. The drop in P@10 of AutoPrompt compared to its P@10 when PLM is fixed (see Table~\ref{tab:result}) may be related to its discrete token substitution (non-gradient-descent) design, which is harder to optimize.

\section{Analysis on Relations with Spurious Associations}
\label{sec: preciseness}
Recent works have shown that frequency bias exists in maximal likelihood estimation training of language models~\cite{pmlr-v80-ott18a, 10.1162/tacl_a_00407} and how a PLM's learning of a word is related to its frequency~\cite{Chang2021WordAI}.~\citet{cao-etal-2021-knowledgeable} observed that prompts for fact probing overfit the object distribution more than the relation. As a result, in factual probing, PLM's output might get affected by the frequencies of output candidates. This is especially true for relations that are prone to spurious associations of the subject with candidate objects or over-representation of candidate objects. Below, we identify some such relations and then analyze performance of SPE with respect to the baselines on these relations. 

\paragraph{R1 Relations with scope associations} (P101 \textit{field of work}). When probing factual knowledge from PLMs, the object of a subject-relation pair forms the correct answer. While there can be multiple reasonable answers, some are more precise and so more desirable than others. For instance, for describing the \textit{field} that \textit{Richard Wagner} worked in (see Table~\ref{tab:example}), both \textit{opera} and \textit{music} seem to be reasonable answers but  \textit{opera} is the more precise one. In such relations, different object-candidates may entail similar meanings but be of different scope. 

\paragraph{R2 Relations with entity-type associations} (P19 \textit{place of birth}, P20 \textit{place of death}, P27 \textit{country of citizenship}, P364 \textit{original language of film or TV show}, P495 \textit{country of origin}, P1412 \textit{language spoken}, P937 \textit{work location}). Some relations are about objects with specific constraints. For example, \textit{place of birth} and \textit{place of death} are the first and last place in a person's life. Those objects, as well as other objects of same entity types that do not match such constraints (e.g. general location names), can co-occur with the subject in the training corpora and get memorized by the pretrained models. Because of these co-occurrences, PLMs may output incorrect objects that are of the correct entity type but may not satisfy the desired constraints. For example, when probed for\textit{ place of birth} of an individual, they may output places where the individual received education or worked instead of where they were born. In the example in Table~\ref{tab:example}, when probing for \textit{citizenship} of famous Brazilian Formula One player \textit{Rubens Barrichello}, P-tuning outputs a handful of countries listed on his Wikipedia page where he participated competitions, which are unrelated to the country of his citizenship, \textit{Brazil}. 

\paragraph{R3 Relations with label distribution associations} (P30 \textit{continent}, P140 \textit{religion}, P413 \textit{position played}, P103 \textit{native language}) \citet{zhong2021factual} showed the label distribution effects prompt-based methods. In particular, for relations with a closed set of candidate objects, the task of factual probing reduces to a classification problem with fixed number of labels. When the correct label (object) appears with very low frequency, PLM's output can get  affected by label distribution in the training set and it can output other labels that appear more frequently. For example, in P30 \textit{continent}, 95.6\% continent-type objects in the training set are \textit{Antartica} (majority class) and only 0.4\% are \textit{Oceania} (minority class). P-tuning is probably affected by this imbalance and outputs the majority label, \textit{Antartica}, as the continent that contains \textit{Marshall Islands} while \textit{Oceania}, the correct answer, appears at rank 4 (see Table~\ref{tab:example}). 

\subsection{Comparison of SPE with Baselines on Relations with Spurious Associations}
\label{sec:preciseness_comparison}
We observe that for R1, R2 and R3 category relations, SPE especially outperformed the baselines in most cases (see Figure~\ref{fig:rels_highlighted}).  The first, second and third rows of the figure represent the corresponding P@1
improvement (scale of 100) of SPE over Optiprompt, SoftPrompt and P-tuning respectively and a darker color means larger improvement. 

\subsection{Investigating Token Frequencies of Top-1 Predictions.}
\label{sec:preciseness_top1}
To further investigate these improvements, we explored the correlation between predictions of different prompt approaches and their token frequencies. We analyzed relations affected by co-occurences of subjects with spurious object candidates, i.e. relations of type R1 and R2. We acquired word frequencies from \citet{robyn_speer_2018_1443582} who collected word frequencies from 8 domains including Wikipedia, books, and news. As discussed in Section~\ref{sec: result} of the paper, we plotted the mean token frequencies of top predictions obtained using different prompting approaches and showed that SPE's predictions  have lower frequencies than most baselines. For example, in the \textit{field of work} of \textit{Richard Wagner} on Table~\ref{tab:example}, the log word frequency of \textit{opera} is -4.73 but the log frequencies of \textit{music}, \textit{history}, \textit{philosophy} and \textit{psychology} are -3.48, -3.61, -4.51 and -4.69 respectively, which have higher word frequencies than \textit{opera} (especially, the frequency of P-tuning's output \textit{music} is 17 times higher). Similarly, in the case of \textit{original language} of \textit{Bazz}, the log frequencies of \textit{Hindi}, \textit{Urdu}, \textit{Punjabi} are -5.18, -5.74, -5.84, while for non-Indian languages like \textit{Turkish}, \textit{English}, and \textit{French} they are -4.71, -3.81 and -3.91, which means these frequencies are at least 10 times higher than the Indian languages. Yet, SPE outputs the correct, even though less frequent answers.

\begin{figure}[t!]
	\centering
	\includegraphics[width=0.5\textwidth]{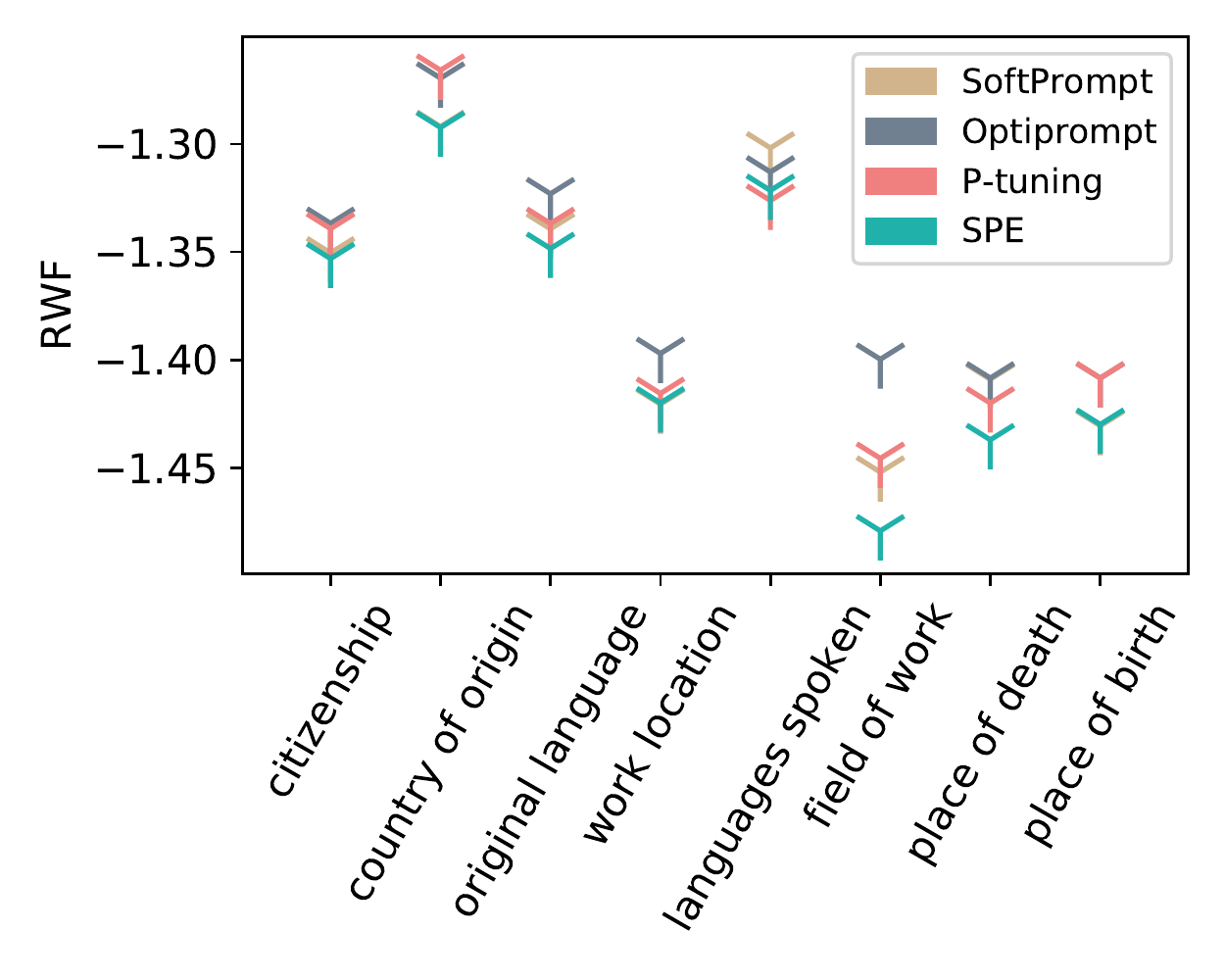}
	\caption{Comparison of different methods using RWF of top 10 predictions for relations with spurious associations. Some markers are not visible because they overlap with the green ones. 
	A lower RWF is better and indicates less association between token frequency and predictions. We see that SPE (green) has lower RWF than P-tuning (pink) in most relations. This indicates that SPE can help PLM in outputing less frequent but correct answers.}
	\label{fig:rank_freq}
\end{figure}

\subsection{Investigating Token Frequencies of Top-M Predictions.}
\label{sec:preciseness_topM}
We now extend the above-mentioned analysis from top predictions to top M predictions and analyze if SPE can help the PLMs output less frequent tokens as answers. In particular, for different prompting approaches, we consider their top M predictions and compute the Rank Weighted Frequency (RWF) using the following formula, where $\mathcal{C}_{n}^{M}$ is the n-th candidates among the top M predictions. 
\begin{gather*}
    \text{RWF} = \sum_{n=1}^{M} \frac{1}{n} \log_{10}(\text{WordFreq}(\mathcal{C}_{n}^{M}))
\end{gather*}

A lower RWF indicates less association between token frequencies and top predictions. Results are shown  in Figure~\ref{fig:rank_freq} with M=10. We can see that for most relations, SPE has a lower RWF than baselines, especially P-tuning. These experiments indicate that SPE can mitigate the frequency bias inherently contained in PLMs and avoid answers with spurious associations with the subjects.

\subsection{Investigating Percentage of Majority Label in Predictions.}
\label{sec:label_dist_association}
\begin{figure}[t!]
	\centering
	\includegraphics[width=0.35\textwidth]{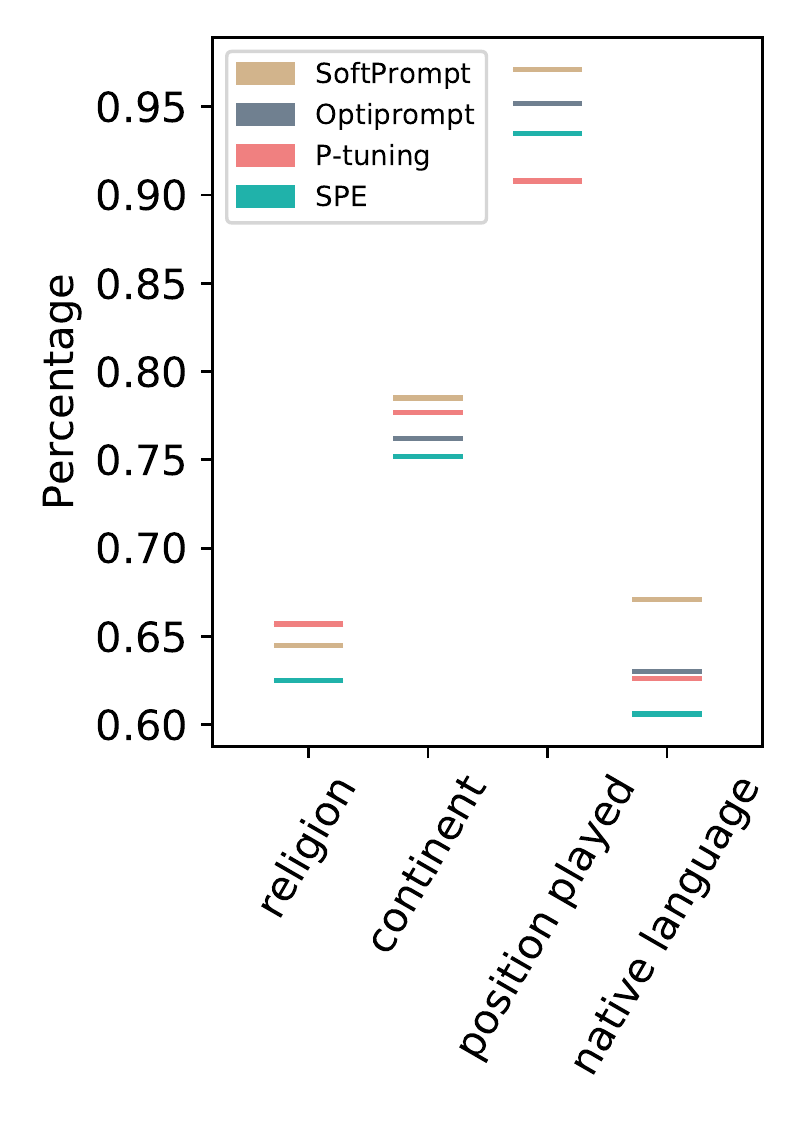}
	\caption{Comparison of different methods in percentage of majority training label in test predictions for R3 relations. A lower value means the method is less affected by the label distribution in the training set. Invisible bars are overlapped with the red ones. SPE (green) has lower values than the baselines in most relations.}
	\label{fig:label distribution asso}
\end{figure}

The analyses shown in Appendix~\ref{sec: preciseness} and \ref{sec:preciseness_topM} focus on relations of type R1 and R2. We now focus on relations of type R3, i.e. relations affected by label imbalance. Results in Figure~\ref{fig:label distribution asso} show that SPE predicts majority training labels less frequently than the baselines in most relations, demonstrating that it is less affected by the imbalances in the label distribution.

\end{document}